\documentclass[lettersize,journal]{IEEEtran}
\usepackage{amsmath,amsfonts}
\usepackage{algorithmic}
\usepackage{algorithm}
\usepackage{array}
\usepackage{textcomp}
\usepackage{stfloats}
\usepackage{url}
\usepackage{verbatim}
\usepackage{graphicx}
\usepackage{cite}
\usepackage{amsthm}
\newtheorem{theorem}{Theorem}
\def\etal{\emph{et al.}}
\def\eg{\emph{e.g.}}
\def\ie{\emph{i.e.}}

\usepackage{subcaption}
\usepackage{wrapfig}
\usepackage{booktabs}
\usepackage{diagbox}
\usepackage{bbding}
\usepackage{makecell}
\usepackage{multirow}
\usepackage{pifont}
\usepackage{bm}
\newcommand{\cmark}{\ding{51}}%
\newcommand{\xmark}{\ding{55}}%

\newif\ifrevise

\revisefalse  % >>> hide revise note <<<

\ifrevise
  \usepackage{makecell}
  \newcounter{mycounter}
  \newenvironment{changedblock}{\color{blue}}{}
  \newcommand{\changed}[1]{\textcolor{blue}{#1}}
  \newcommand{\changedtab}[1]{\color{blue}{#1}}
  \newcommand{\mycell}[1]{
    \makecell[c]{
      \textbf{{\Large \arabic{mycounter}}} \\ \textbf{{\scriptsize #1}}
    }
  }
  \newcommand{\margintextnote}[1]{
    \stepcounter{mycounter}
    \marginpar[\textcolor{blue}{\mycell{#1}}]{\textcolor{blue}{\mycell{#1}}}
  }
  \newcommand{\intextnote}[1]{\textcolor{blue}{(\textbf{\normalsize{#1}}) }}
\else
  \newenvironment{changedblock}{\color{black}}{}
  \newcommand{\changed}[1]{\textcolor{black}{#1}}
  \newcommand{\changedtab}[1]{\color{black}{#1}}
  \newcommand{\margintextnote}[1]{}
  \newcommand{\intextnote}[1]{}
\fi
\usepackage[pagebackref=true,breaklinks=true,letterpaper=true,colorlinks,bookmarks=false]{hyperref}

\begin{document}

\title{ScaleNet: Scaling up Pretrained Neural Networks \\ with Incremental Parameters}

\author{
    Zhiwei~Hao, Jianyuan~Guo, Li~Shen, Kai~Han, Yehui~Tang, Han~Hu, and~Yunhe~Wang
    % <-this % stops a space
    % \thanks{Manuscript received xx xx, 202x; revised xx xx, 202x.}
    \thanks{Corresponding author: Yunhe~Wang.}
    \thanks{This work was supported by the Joint Funds of the National Natural Science Foundation of China (NSFC) under Grant No. U2336211, by the NSFC under Grant No. 62576364, and in part by the Start-up Grant (No. 9382010) of the City University of Hong Kong.}
    \thanks{Zhiwei~Hao and Han~Hu are with the School of Information and Electronics, Beijing Institute of Technology, Beijing 100081, China (e-mail: haozhw@bit.edu.cn, hhu@bit.edu.cn).}
    \thanks{Jianyuan~Guo is with the Department of Computer Science, City University of Hong Kong, HKSAR, China. (e-mail: jianyguo@cityu.edu.hk).}
    \thanks{Li~Shen is with the School of Cyber Science and Technology, Shenzhen Campus of Sun Yat-sen University, Shenzhen 510275, China (e-mail: mathshenli@gmail.com).}
    \thanks{Kai~Han, Yehui~Tang, and~Yunhe~Wang are with Huawei Noah's Ark Lab, Beijing 100084, China (e-mail: kai.han@huawei.com, yehui.tang@huawei.com, yunhe.wang@huawei.com).}
}

\markboth{}%
{Hao \MakeLowercase{\textit{et al.}}: ScaleNet}


\maketitle

\begin{abstract}
Recent advancements in vision transformers (ViTs) have demonstrated that larger models often achieve superior performance. However, training these models remains computationally intensive and costly. To address this challenge, we introduce ScaleNet, an efficient approach for scaling ViT models. Unlike conventional training from scratch, ScaleNet facilitates rapid model expansion with negligible increases in parameters, building on existing pretrained models. This offers a cost-effective solution for scaling up ViTs.
Specifically, ScaleNet achieves model expansion by inserting additional layers into pretrained ViTs, utilizing layer-wise weight sharing to maintain parameters efficiency. Each added layer shares its parameter tensor with a corresponding layer from the pretrained model. To mitigate potential performance degradation due to shared weights, ScaleNet introduces a small set of adjustment parameters for each layer. These adjustment parameters are implemented through parallel adapter modules, ensuring that each instance of the shared parameter tensor remains distinct and optimized for its specific function.
Experiments on the ImageNet-1K dataset demonstrate that ScaleNet enables efficient expansion of ViT models. With a 2$\times$ depth-scaled DeiT-Base model, ScaleNet achieves a 7.42\% accuracy improvement over training from scratch while requiring only one-third of the training epochs, highlighting its efficiency in scaling ViTs. Beyond image classification, our method shows significant potential for application in downstream vision areas, as evidenced by the validation in object detection task.
\end{abstract}


\section{Introduction}
\IEEEPARstart{R}{ecent} advances in the computer vision community have highlighted the effectiveness of scaling up vision transformers (ViTs)~\cite{vit,swin,guo2022cmt,guo2022hire}.
Central to these models is the principle that ``larger is better,'' as performance consistently improves with increased model size. This is often referred to as the scaling law~\cite{kaplan2020scaling}. Due to its simplicity and effectiveness, an increasing number of researchers and companies are opting to scale up their models to enhance performance.

Scaling up model sizes leads to remarkable performance gains, but it comes at a substantial cost. Take the ``huge'' variant of the vanilla ViT architecture as an example: it contains 632 million parameters and requires 167 billion FLOPs for each forward pass. While this model significantly outperforms the base version, its parameter count and computational complexity are approximately 7.3 and 9.5 times larger, respectively~\cite{vit}. Similar trends are also observed in the natural language processing domain~\cite{llama3,hao2025adem}. This surge in computational demands not only presents serious environmental challenges, particularly with regard to carbon emissions, but also limits access for researchers with constrained computational resources. Additionally, the increased parameter count leads to substantial storage requirements. To address these issues, there is an urgent need for methods that can scale models rapidly while maintaining parameter efficiency.

One approach to model scaling is progressive training, where a larger model is partially or fully initialized using parameters from a smaller, pretrained model. In earlier works, parameters from the pretrained model are directly transferred to a subset of the larger model, with the remaining parameters being randomly initialized~\cite{net2net}. During training, the pretrained subset can either remain fixed or be finetuned. In the context of transformer models, newly introduced parameters are typically derived from the parameters in pretrained models. These approaches mainly focus on two aspects: width growth and depth growth. Width growth involves expanding layer dimensions by duplicating parameters~\cite{shen2022staged, compoundgrow} or applying complex mapping rules~\cite{bert2bert, ligo}, while depth growth increases depth of the model by duplicating and stacking layers~\cite{li2020shallow, gong2019efficient, mogrow}. Some works combine both techniques~\cite{ligo, shen2022staged, compoundgrow, bert2bert}. Progressive training approaches exploit prior knowledge from smaller pretrained models, facilitating faster training of scaled models.

\IEEEpubidadjcol

Although progressive training efficiently scales models by increasing parameters proportionally, it does not offer benefits in terms of parameter efficiency. Moreover, the addition of numerous free parameters potentially slow down the optimization process due to the expanded parameter space. However, recent advancements in compact models demonstrate that over-parameterized models can learn generalizable knowledge through their weights, enabling controlled weight sharing without performance loss~\cite{savarese2019learning,minivit}. By incorporating identical shared parameters from the pretrained model to build the scaled model, we effectively shrink the optimization landscape. This not only speeds up the optimization process but also significantly enhances parameter efficiency.

\begin{figure}
    \centering
    \includegraphics[width=0.95\linewidth]{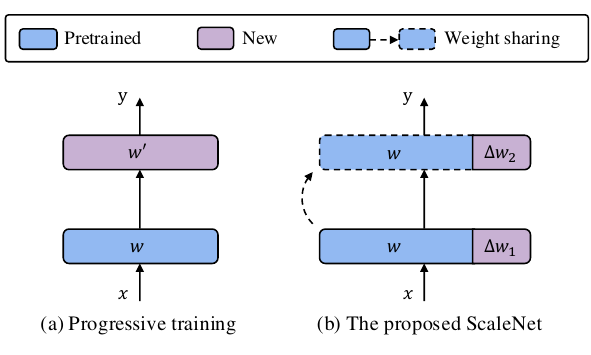}
    \caption{Comparison between (a) progressive training and (b) the proposed ScaleNet method. Progressive training initializes new layers with pretrained layers, and each layer is treated as independent during subsequent training. In contrast, ScaleNet shares the same parameters between the new layers and the pretrained layers, with only a small fraction of adjustment parameters to achieve layer-specific functions, surpassing progressive training in terms of parameter efficiency.}
    \label{fig:cover}
\end{figure}

To this end, we introduce ScaleNet, a method for expanding pretrained ViTs, which applies weight sharing to ensure both training and parameter efficiency. We employ a layer-wise expansion scheme, where the scaled model is obtained by inserting additional layers to increase the depth of pretrained ViTs. Weight sharing is performed at the layer level, with each added layer sharing its parameter tensor with the corresponding layer from the pretrained model. Identical weight sharing could potentially lead to model collapse. Drawing inspiration from recent advancements in parameter-efficient fine-tuning (PEFT), we introduce a small set of adjustment parameters for each shared layer through lightweight adapter modules. These modules introduce additional information into the linear layers of each multi-layer perceptron (MLP) module, ensuring that each instance of the shared parameter tensor remains distinct.
Figure~\ref{fig:cover} presents a comparison between progressive training and our ScaleNet, where the adoption of weight sharing and layer-specific adjustment parameters avoids the significant increase in parameters caused by the introduction of new layers.
To validate the effectiveness of the proposed ScaleNet method, we conduct experiments on the ImageNet-1K dataset~\cite{imagenet} using two typical ViT architectures: DeiT~\cite{deit} and Swin~\cite{swin}. Our experimental results demonstrate that ScaleNet can scale pretrained ViT models efficiently.
Our main contributions can be summarized as follows:
\begin{itemize}
    \item We propose ScaleNet, a novel method for expanding pretrained Vision Transformers (ViTs) that employs weight sharing to improve both training and parameter efficiency.
    \item We introduce a layer-wise expansion scheme that scales models efficiently by inserting additional layers and sharing parameters at the layer level, avoiding model collapse through lightweight adapter modules.
    \item We conduct extensive experiments on the ImageNet-1K dataset, where results demonstrate that ScaleNet significantly outperforms training a scaled model from scratch, achieving a 7.42\% accuracy improvement while requiring only one-third of the training epochs on a 2$\times$ depth-scaled DeiT-Base model.
\end{itemize}

\section{Related work}

\subsection{Progressive Training}

Progressive training initially focused on mitigating the training challenges of deep neural networks~\cite{bengio2006greedy,fahlman1989cascade,hinton2006fast}. Today, it primarily addresses the escalating computational costs associated with modern deep learning models. Early efforts, such as Net2Net~\cite{net2net}, pioneered techniques to accelerate the training of large models by leveraging initialization from smaller, pretrained models. These approaches mainly involved expanding model width by duplicating parameters or increasing depth by stacking layers.

In natural language processing (NLP), progressive training has been particularly impactful in optimizing the pretraining of models like BERT~\cite{bert}. Researchers have employed strategies such as duplicating parameters~\cite{shen2022staged, compoundgrow} or applying complex mapping rules~\cite{bert2bert, ligo} to expand layer dimensions, while some works increase the depth of the model by duplicating and stacking layers~\cite{li2020shallow, gong2019efficient}. These approaches not only enable faster training but also ensure that scaled models benefit from the prior knowledge embedded in their smaller counterparts. Similar concepts have also been explored in computer vision~\cite{mogrow,ligo}.

Although progressive training accelerates the training of large models, it does not offer improvements in parameter efficiency. Furthermore, the increase in trainable parameters may potentially slow down the optimization process due to the expanded parameter space.

\subsection{Weight Sharing}

Weight sharing involves reusing subsets of parameters within a model across two or more modules, with the goal of reducing the total number of parameters that need to be stored. Despite being proposed decades ago~\cite{lecun1989generalization}, the concept of weight sharing has persisted and found applications across various fields in the modern era of deep learning.

The initial attempts at weight sharing occurred in the NLP field. Lan \etal~\cite{albert} proposed sharing all model parameters across layers, achieving superior performance to BERT-large~\cite{bert} with significantly fewer parameters. Furthermore, subsequent works have adopted weight sharing for various NLP tasks~\cite{dehghani2018universal,bai2019deep,savarese2019learning,ma2019tensorized,reid2021subformer,liu2023enhancing}. Recently, weight sharing has been extended to vision tasks. 
\changed{\margintextnote{R1-Ref}In the multi-task learning context, Sun~\etal~\cite{sun2020learning} introduced a sparse sharing mechanism, where each task extracts a task-specific subnetwork from a large, over-parameterized base network.}
Zhang \etal~\cite{minivit} proposed weight sharing across blocks in vision transformers, significantly reducing the parameter count without compromising performance. Existing weight sharing approaches primarily focus on building compact models by reducing parameter count, whereas our method focuses on model expansion, distinguishing it from these approaches.

\subsection{Parameter Efficient Fine-tuning}

PEFT is a technique for adapting a pretrained model to downstream tasks by modifying only a small subset of its parameters. By constraining the number of trainable parameters, PEFT not only enhances memory efficiency but also accelerates training speed. In some cases, PEFT has also showcased superior performance compared to full fine-tuning~\cite{lialin2023scaling}.

Existing PEFT approaches can be roughly classified into selective, prompt-based, adapter-based, and reparameterization-based methods.
Selective approaches adopt sparse fine-tuning of pretrained models~\cite{diffpruning,fishmask,zhang2023gradient}. For instance, BitFit~\cite{bitfit} exclusively fine-tunes the biases of the pretrained model while keeping all other parameters fixed, achieving comparable performance to full fine-tuning. Prompt-based methods control model performance by modifying the input and are frequently customized for LLM, utilizing either discrete spaces~\cite{gpt3,schick2020s} or continuous spaces~\cite{lester2021power,li2021prefix}. Furthermore, there are also works designing prompts for vision tasks~\cite{vpt,lvm,guo2024dataefficient}. The original adapter method introduces additional trainable linear layers in each transformer block. Currently, the idea of introducing additional trainable parameters has been extended by using more complex adapter architectures. The original adapter method introduced additional trainable linear layers within each transformer block~\cite{adapter}. Presently, this idea has been expanded by including more complex adapter architectures~\cite{pfeiffer2020adapterhub,vitadapter,wang2024hada}. Reparameterization-based approaches share similarities with adapter-based methods in introducing new trainable parameters. However, they differ in that the additional parameters can be merged into the original model without incurring any additional inference cost. One of the most prominent methods in this class is LoRA~\cite{lora}, which decomposes the parameter update of a weight matrix into the product of two low-rank matrices. This approach has demonstrated effectiveness in various scenarios. 
\changed{\margintextnote{R3-Q3}Crucially, ScaleNet addresses the challenge of model expansion, a distinct goal from the task adaptation targeted by PEFT methods. While PEFT methods learn small weight modifications to adapt a fixed-size model to a new task, ScaleNet fundamentally changes the model architecture by adding new layers to enhance its performance, often on the same task.
}
\section{Method}

\subsection{Preliminaries}

\subsubsection{Vision Transformer}

Vision transformers~\cite{vit} are attention-based models inherited from NLP, where each layer consists of a multi-head self-attention (MSA) block and a multi-layer perceptron (MLP) block. These models take image patches as input, where the original input image is first reshaped into flattened patches. The patches are then projected into an embedding space and added to positional embeddings. 
Suppose the embeddings are denoted as $\bm{x}$, the workflow of one layer in ViT is as follows:
\begin{align}
  \bm{x} &\gets \text{MSA}(\text{LN}(\bm{x})) + \bm{x}, \\
  \bm{x} &\gets \text{MLP}(\text{LN}(\bm{x})) + \bm{x},
\end{align}
where LN denotes layer normalization~\cite{ba2016layer}.

\subsubsection{Progressive Training}

Progressive training accelerates the training of large models by leveraging initialization from smaller, pretrained models. This can be achieved either by expanding the width through larger feature dimensions or by increasing the depth by adding model layers for feature extraction. 

An example of a width-expanding method is Net2Net~\cite{net2net}. Given a pretrained parameter tensor $\bm{\theta} \in \mathbb{R}^{n \times n}$, Net2Net expands it into a larger tensor $\bm{\theta'} \in \mathbb{R}^{m \times m}$, where $m > n$, by duplicating the rows and columns of $\bm{\theta}$. Specifically, $\bm{\theta}$ is placed in the upper-left corner of $\bm{\theta'}$, and the remaining columns of $\bm{\theta'}$ are filled by randomly selecting from $\bm{\theta}$ with replacement. The rows are filled similarly, with an additional normalization step to ensure that all neurons receive the same input values as those in the original network.

Depth-expanding methods include approaches like StackBERT~\cite{gong2019efficient}, which doubles the depth of a pretrained model by adding new layers and initializing them with the weights from the corresponding pretrained layers. Another method is interpolation~\cite{chang2017multi,dong2020towards}, where new layers are interleaved into the pretrained model and initialized using weights from adjacent pretrained layers. To expand a model from $L$ layers to $L' = nL$ layers, the newly added layers are initialized as:
\begin{equation}
    \begin{aligned}
        \bm{\theta'_l} &\gets \bm{\theta}_{l \mod L}, & \text{(StackBERT)} \\
        \bm{\theta'_l} &\gets \bm{\theta}_{\lfloor l/n \rfloor}. & \text{(Interpolation)}
    \end{aligned}
    \label{eq:baseline}
\end{equation}
where $l$ is the index of each layer. In this paper, we focus on depth-wise expansion for model scaling, without considering width expansion.

\subsubsection{Weight Sharing}

Weight sharing is a technique designed to reduce model size while maintaining performance by reusing subsets of parameters across two or more layers within a model. For a group of consecutive layers $l \in \mathcal{L}$ sharing the same weight $\bm{\theta_{\mathcal{L}}}$, their forward pass is formulated as:
\begin{equation}
    \bm{x}_{l+1} = f(\bm{x}_l; \bm{\theta_{\mathcal{L}}}),
    \label{eq:ws}
\end{equation}
where $\bm{x}_l$ is the input of layer $l$. By reusing the same weight matrix across multiple layers, the total number of parameters that need to be stored is reduced. This reduction in independent parameters also shrinks the optimization landscape, which can potentially accelerate the optimization process.

\subsection{Scaling up of Pretrained ViTs}

\begin{figure*}
    \centering
    \includegraphics[width=0.95\linewidth]{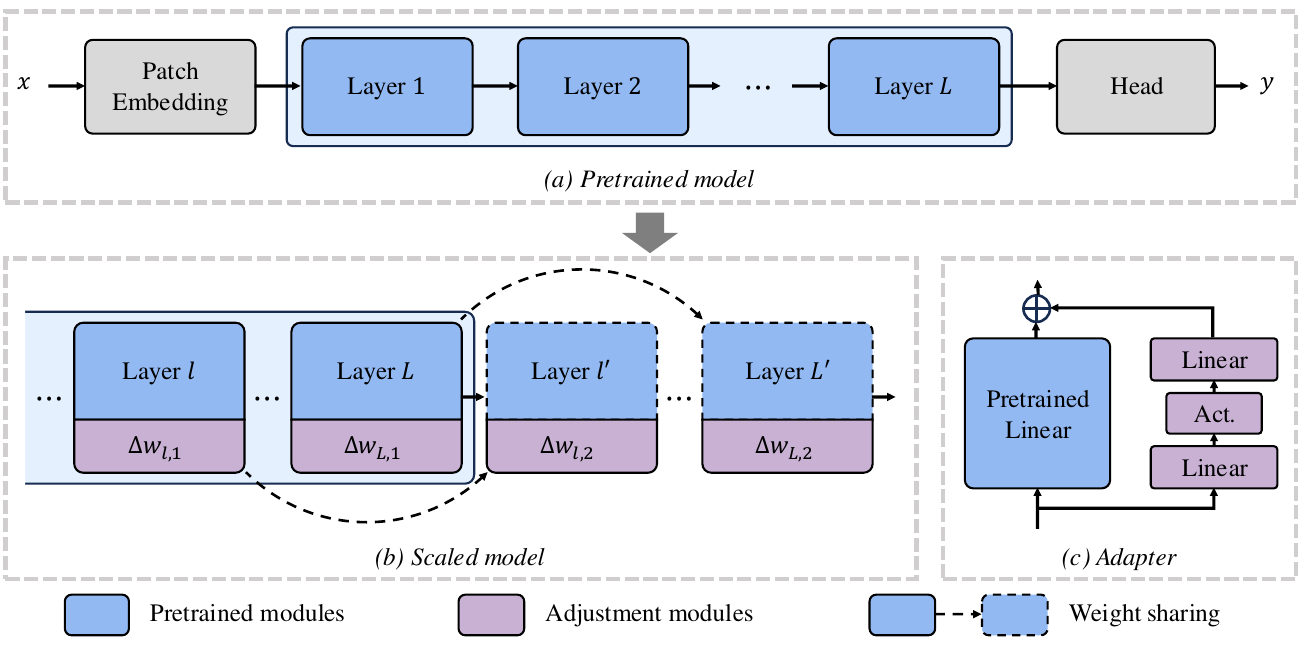}
    \caption{Overview of the ScaleNet method. (a) A pretrained ViT model consists of a patch embedding layer, intermediate layers, and a prediction head, where the intermediate layers are the target for model scaling. (b) Scaling the pretrained $L$-layer model to $L'$ layers. In the figure example, new layers are added at the tail of the pretrained layers, with each new layer sharing weights with a corresponding pretrained layer. Other layer mapping strategies are explored in our experiments. Layer-specific adjustment modules are adopted to differentiate the weight-shared layers, improving model capacity. (c) Parallel adapter-based adjustment module, which consists of a down-projection layer, an up-projection layer, and a non-linear activation function. This module is placed at all linear layers in MLP modules of the ViT.}
    \label{fig:main}
\end{figure*}

To address the challenge of the high computational costs associated with training large ViT models, we propose a method called ScaleNet, which preserves the advantages of both training and parameter efficiency. 
Figure~\ref{fig:main} presents an overview of the ScaleNet framework, where a pretrained $L$-layer model is scaled to $L'$ layers. The scaling is achieved through weight-sharing-based progressive training, \ie, each newly added layer share weights with a corresponding layers in the pretrained model. Moreover, parallel adapters are adopted at all linear layers in the MLP modules to increase diversity in layers sharing the same weights, which helps increase the capacity of the scaled model while maintaining parameter efficiency.
We first introduce our implementation of weight sharing-based progressive training for ViT models.

\subsubsection{Depth-wise Model Scaling}

Suppose a pretrained model consists of $L$ layers, with the weight of the $l$-th layer denoted as $\bm{\theta}_l$. When adopting depth-wise scaling to build a model with $L'$ layers, where $L' > L$, the $l'$-th layer in the scaled model is initialized by:
\begin{equation}
    \bm{\theta}_{l'} \gets \bm{\theta}_{g(l')},
\end{equation}
\begin{changedblock}
where $g(l')$ defines the layer mapping between the scaled model with layer index $l' \in \{1, \dots, L'\}$ and the pretrained model with layer index $l \in \{1, \dots, L\}$. For layers that are part of the original model ($l' \le L$), the mapping is an identity function, i.e., $g(l')=l'$. For newly added layers ($l' > L$), various mapping strategies can be employed. A common and effective strategy, inspired by methods like StackBERT, is a cyclic mapping, defined as:\margintextnote{R1-Q3}
\begin{equation}
g(l') =
\begin{cases}
l', & \text{if } l' \leq L, \\
l' \mod L , & \text{if } l' > L.
\end{cases}
\label{eq:mapping}
\end{equation}
This function maps each new layer $l'$ to a corresponding layer in the original model $\{1, 2, \dots, L\}$ in a cyclic manner. For instance, with this rule, layer $L+1$ would be mapped to layer 1, layer $L+2$ to layer 2, and so on, repeating the pattern of the pretrained layers. While this cyclic mapping is a primary example, other mapping functions are also possible. We explore the effects of different strategies in our experiments. Although the newly added layers in the scaled model are initialized with corresponding layers from the pretrained model, they remain independent during training in conventional progressive training. This independence results in higher training and inference costs, as well as increased storage space requirements.
\end{changedblock}

Drawing inspiration from weight sharing, we require the newly added layers to share weight tensors with layers in the original pretrained model, following the layer mapping defined by $g(l')$ in Equation~\ref{eq:mapping}. This mapping not only defines the initialization of the corresponding layers but also ensures that the mapped layers share the same weights during the subsequent training period. As a result, expanding the model depth does not lead to a proportional increase in model parameters, ensuring parameter-efficient model scaling.

\subsubsection{Layer Adjustment}

Weight sharing was originally designed for training lightweight models from scratch. In the model expansion scenario, however, training begins with model layers already adapted to the pretraining dataset. While weight sharing significantly improve parameter efficiency in the scaled model, it may negatively impact performance due to the limited learning capacity of pretrained parameters.

To bridge this gap, we modify each layer to enhance its model capacity, ensuring that each instance of the shared parameter tensor learns distinct functions, thereby improving overall model performance. These modifications are implemented by introducing adjustment parameters for each layer. By incorporating lightweight adjustment modules, the parameter efficiency of the scaled model is preserved. Specifically, let the adjustment parameters of layers $l \in \mathcal{L}$ be denoted as $\Delta \bm{\theta}_{l}$, then the workflow in Equation~\ref{eq:ws} becomes:
\begin{equation}
    \bm{x}_{l+1} = f(\bm{x}_l; \bm{\theta}_{\mathcal{L}}, \Delta \bm{\theta}_{l}).
\end{equation}
This adjustments can be implemented using the following two methods.

\textbf{LoRA.}
Low-rank adaptation (LoRA) is a method to finetune pretrained models on downstream tasks parameter-efficiently~\cite{lora}.
It reduces the number of trainable parameters by learning pairs of rank-decomposed matrices while keeping the original weights frozen, which is formulated as follows:
\begin{equation}
    f(\bm{x}) = (\bm{\theta} + \bm{BA})\bm{x},
\end{equation}
where $\bm{B}\in\mathbb{R}^{d\times r}$ and $\bm{A}\in\mathbb{R}^{r\times k}$, with the rank $r \ll \min(d,k)$ are the low-rank decomposition matrices.
$\bm{\theta}\in\mathbb{R}^{d\times k}$ indicates the pretrained weights.
Matrix $\bm{A}$ is Gaussian initialized, and $\bm{B}$ is zero-initialized to ensure $\bm{BA}$ is zero at the beginning of training.
To achieve layer adjustment, we assign each instance in a group of layers sharing the same weight a unique LoRA module.

\textbf{Adapter.}
Adapters fine-tune pretrained models by inserting additional linear layers into them. These linear layers can be placed at various positions within the model. Similar as LoRA, we adopt the design of a parallel adapter~\cite{he2021towards}, which operates as follows:
\begin{equation}
    f(\bm{x}) = \bm{\theta}\bm{x} + \bm{B}\sigma(\bm{A}\bm{x}),
\end{equation}
where $ \sigma $ is a non-linear activation function.
Note that this parallel adapter can be regarded as an extended version of LoRA, which incorporates an additional non-linear operation for improved representation capacity. When using LoRA as an adjustment module, its parameters cannot be merged into the original layer because of the use of weight sharing. Therefore, the parallel adapter does not sacrifice parameter efficiency compared to LoRA. 
The left bottom of Figure~\ref{fig:main} presents an illustration of the parallel adapter architecture.
In our experiments, we will compare the adoption of LoRA and the parallel adapter as adjustment modules.

\subsection{Training of Scaled Model}

After constructing the scaled model, a training process is required to enhance the cooperation among the layers and adjustment parameters. Furthermore, to facilitate better learning of the scaled model, we keep the normalization layers independent, \ie, these layers are only initialized through progressive training, without adopting weight sharing. Since the parameter count in normalization layers is minimal, this does not have a significant impact on parameter efficiency.
Hence, the training objective is:
\begin{equation}
    \max_{\{\Delta \bm{\theta}_{l}| l \in \mathcal{L}\}}\sum_{(\bm{x},y)\in \mathcal{Z}}\log(\bm{P}_{\bm{\theta}_\mathcal{L};\Delta \bm{\theta}_{l}}(y|\bm{x})).
\end{equation}
As depicted in the training objective, only the adjustment parameters $ \Delta \bm{\theta}_{l} $ are optimized. This configuration is inspired by attempts to fine-tune LLMs, where a large pretrained model can adapt to various downstream tasks by tuning only its parameter-efficient modules. However, for smaller models such as DeiT-Tiny and DeiT-Small~\cite{deit}, we allow all model parameters to be updated to facilitate their learning.

\subsection{Discussion}

By adopting weight-sharing and lightweight adjustment modules, the expanded model obtained by ScaleNet remains parameter-efficient. Suppose a pretrained model consists of $L$ layers, with each layer containing $n$ linear layers and an intermediate dimension of $d$. Under a 2-times model scaling scenario, let the intermediate dimension of the adjustment module be $r$. The scaled model contains only a fraction of $\frac{Lnd^2 + 4Lnrd}{2Lnd^2}$ of the parameters compared to the scaled model in existing progressive training methods. Since $r \ll d$ typically holds, this fraction approximates to $1/2$.

Although our scaled model maintains a similar parameter count, its computational complexity is comparable to that of the expanded model in existing progressive training methods. The increase in computational operations is akin to recent advancements in test-time computing~\cite{ji2025test}, where model performance is enhanced by increasing test-time computation without a corresponding increase in model parameters. In this way, ScaleNet also facilitates the deployment of powerful models on edge devices, where improved performance can be achieved without a significant increase in memory requirements.

\begin{changedblock}

\subsection{Theoretical Analysis}
\margintextnote{R2-Q2}

This section aims to provide a theoretical foundation for the effectiveness of ScaleNet. From the perspective of Approximation Theory, we will demonstrate that the ScaleNet architecture possesses an efficient Universal Approximation Capability.

\subsubsection{Problem Setting}

We first formalize the ScaleNet architecture. Let the input image be divided into $N$ patches, each of which is linearly projected into a $d$-dimensional vector. The input can thus be represented as a matrix $\bm{x} = [x_1, \dots, x_N] \in [0, 1]^{d \times N}$, where we assume the input vectors have been normalized to the compact set $[0, 1]^d$.

The core of ScaleNet is a recurrent structure. A single recurrent step consists of a shared, fixed-parameter Transformer layer $T: \mathbb{R}^{d \times N} \to \mathbb{R}^{d \times N}$ and an independent, trainable, lightweight Adapter module $A^{(t)}$. Parameters of adapters are not shared across different recurrent steps $t$. If the total number of recurrent steps is $n$, the evolution of the hidden state $H^{(t)}$ is defined as:
\begin{equation}
H^{(t)} = T\left(H^{(t-1)}\right) + A^{(t)}\left(H^{(t-1)}\right), \quad t = 1, \dots, n,
\end{equation}
with the initial state $H^{(0)} = \bm{x}$. The final output $Y \in \mathbb{R}^C$ is obtained by applying a classification head, $\text{Head}$, to the final hidden state $H^{(n)}$.

Our goal is to prove that the function $\hat{f}: [0, 1]^{d \times N} \to \mathbb{R}^C$ defined by this architecture can approximate any continuous target function $f^*$. To quantify the smoothness of the function, we introduce its modulus of continuity, defined as $\omega_{f^*}(\delta) := \sup_{\|\bm{x} - \bm{x}'\|_F \le \delta} \|f^*(\bm{x}) - f^*(\bm{x}')\|_\infty$.

\subsubsection{Universal Approximation Capability of ScaleNet}

The following theorem quantitatively establishes the relationship between the approximation capability of ScaleNet and its effective depth.

\begin{theorem}[]
For any continuous function $f^*$ on the compact set $[0,1]^{d\times N}$, and for any given approximation accuracy $\varepsilon > 0$, there exist a sufficiently fine partition granularity $\delta > 0$ and a sufficiently large number of recurrent steps $n = \mathcal{O}\left(dN \log(\frac{1}{\delta})\right)$, such that we can construct a ScaleNet model $\hat{f}$ by training the adapters to satisfy:
\begin{equation}
\sup_{\bm{x} \in [0,1]^{d \times N}} \left\| \hat{f}(\bm{x}) - f^*(\bm{x}) \right\|_\infty \le C_1  \omega_{f^*}(\delta) + C_2  \delta,
\end{equation}
where $C_1$ and $C_2$ are constants that depend only on the fixed model dimensions $(d, N, C)$. By choosing a sufficiently small $\delta$, the total error can be made smaller than any $\varepsilon$.
\label{theorem}
\end{theorem}

\subsubsection{Proof}
Our proof is constructive and consists of three main steps inspired by~\cite{xu2024expressive}. First, we approximate the continuous target function $f^*$ with a piecewise constant function $\bar{f}$. Second, we show how the first Adapter $A^{(1)}$ can be used to perform parallel quantization on all input patches to identify the partition region they belong to. Finally, we explain how the subsequent recurrent structure implements the mapping from the quantized ID to the target value.

\textbf{Step 1: Approximating the target function with a piecewise constant function.}
Any continuous function on a compact set can be approximated arbitrarily well by a piecewise constant function. We partition the input space for each patch, $[0,1]^d$, into $\delta^{-d}$ non-overlapping $d$-dimensional hypercubes by dividing each dimension into $\delta^{-1}$ intervals of length $\delta$. For the entire input space $[0,1]^{d \times N}$, this corresponds to a partition of $M = (\delta^{-d})^N = \delta^{-dN}$ combined hypercubes.

We assign a representative point $\bm{x}_B$, \eg, the center, to each combined hypercube and define the piecewise constant function $\bar{f}(\bm{x}) := f^*(\bm{x}_B)$, where $\bm{x}$ and $\bm{x}_B$ belong to the same hypercube. By the uniform continuity of $f^*$ and the definition of the modulus of continuity, the approximation error is:
\begin{equation}
\|f^*(\bm{x}) - \bar{f}(\bm{x})\|_\infty \le \omega_{f^*}(\|\bm{x} - \bm{x}_B\|_F) \le \omega_{f^*}(\sqrt{Nd}\delta).
\end{equation}
Since $N$ and $d$ are constants, this error term can be denoted as $\mathcal{O}(\omega_{f^*}(\delta))$.

\textbf{Step 2: Implementing parallel quantization with the first Adapter.}
To implement the function $\bar{f}$, the model must first be able to identify the combined hypercube to which an arbitrary input $\bm{x}$ belongs. This is equivalent to determining the ID of each patch $x_n$. We leverage the first Adapter $A^{(1)}$ for this task, which takes the initial state $H^{(0)} = \bm{x}$ as its input.

According to the Universal Approximation Theorem, a single-hidden-layer ReLU network $A^{(1)}$ can be constructed to process each patch $x_n$ in parallel and map it to an encoding that uniquely identifies the $\delta$-hypercube it belongs to. Thus, the output of $A^{(1)}(\bm{x})$ contains the complete quantization information. The subsequent state $H^{(1)} = T(\bm{x}) + A^{(1)}(\bm{x})$ then serves as the input for the value mapping stage, carrying both the features transformed by $T$ and the precise quantization IDs provided by $A^{(1)}$. The approximation error introduced by this quantization step can be controlled to be $\mathcal{O}(\delta)$.

\textbf{Step 3: Implementing value mapping with the recurrent structure.}
The model then needs to map one of the $M = \delta^{-dN}$ possible ID combinations, encoded in $H^{(1)}$, to its corresponding function value $\bar{f}(\bm{x}_B)$. This process is accomplished by the subsequent $n-1$ recurrent steps, which collectively form a deep network that essentially implements a large lookup table.

The partitioning capability of a deep ReLU network grows exponentially with its depth, \ie, the number of recurrent steps $n$. To implement a lookup table function with $M$ pieces, the required network depth is proportional to $\log M$ \cite{montufar2014number,telgarsky2016benefits}. Therefore, the required number of recurrent steps $n-1$ satisfies:
\begin{equation}
    n-1 = \mathcal{O}(\log M) = \mathcal{O}\left(dN \log\left(\frac{1}{\delta}\right)\right).
\end{equation}
The approximation error introduced in this step can also be controlled to be $\mathcal{O}(\delta)$.

\noindent\textbf{Error Summary.}
Combining the errors from all steps, we obtain an upper bound on the total approximation error of ScaleNet:
\begin{equation}
    \begin{aligned}
        \sup_{\bm{x}} \|\hat{f}(\bm{x}) - f^*(\bm{x})\|_\infty &\le \|\hat{f}(\bm{x}) - \bar{f}(\bm{x})\|_\infty\\
        &+ \|\bar{f}(\bm{x}) - f^*(\bm{x})\|_\infty\\
        &\le \mathcal{O}(\delta) + \mathcal{O}(\omega_{f^*}(\delta))\\
        &= C_1 \omega_{f^*}(\delta) + C_2 \delta.
    \end{aligned}
\end{equation}
For any given $\varepsilon > 0$, since $\omega_{f^*}(\delta) \to 0$ as $\delta \to 0$, we can always choose a sufficiently small $\delta$ such that $C_1 \omega_{f^*}(\delta) + C_2 \delta < \varepsilon$. This requires the corresponding number of recurrent steps $n$ to be sufficiently large, with its magnitude given by Theorem~\ref{theorem}. This completes the proof.

\noindent\textbf{Conclusion.} This theoretical analysis validates the effectiveness of the ScaleNet architecture. Through the weight-sharing mechanism, ScaleNet converts limited parameter increments into substantial increases in network depth. This allows it to expand its capacity and functional complexity more efficiently and fundamentally than PEFT methods, leading to superior performance in model scaling. 

\end{changedblock}

\section{Experiment}

\subsection{Implementation Details}

\subsubsection{Model}

To evaluate the performance of our proposed ScaleNet in scaling pretrained ViT models, we employ two typical ViT architectures: the isotropic DeiT~\cite{deit} model and the hierarchical Swin~\cite{swin} model for our experiments. The pretrained models are obtained from their officially released pretrained checkpoints. By default, we scale all layers in DeiT architectures and scale the last stage in Swin architectures, both with a scaling factor of 2, \ie, each scaled pretrained layer shares its weight with only one newly added layer. The new layers of DeiT models are interpolated into the model, while those of Swin models are inserted at the tail of the pretrained model. The intermediate dimension of all the adjustment modules is set to 16. We will ablate other model scaling configurations in our experiments. For DeiT-Tiny and DeiT-Small, all parameters in the model are trainable, whereas for all other models, only the parameters in the adjustment modules and layer normalization are trainable. This choice is based on the consideration that larger models learn general representations, so tuning only a small fraction of the model is sufficient for adapting the scaled model, which also helps achieve faster convergence.

\begin{table}
    \renewcommand\tabcolsep{6pt}
    \centering
    \caption{\changed{\intextnote{R1-Q2/R2-Q1}Performance comparison of ScaleNet with Stack and Interpolate baselines on ImageNet. The table reports Top-1 accuracy and the number of parameters for each model, showing that ScaleNet achieves higher accuracy with fewer parameters compared to the baseline methods.}}
    \changedtab{\begin{tabular}{l|c|c|c|c}
        \toprule
        Model                                   & Expand      & \#Params (M) & FLOPs (G) & Top-1 \\
        \midrule
        \multirow{6}{*}{Deit-Tiny~\cite{deit}}  & -           & 5.72     & 1.3 & 72.16      \\
                                                & Random      & 6.61     & 1.5 &  75.01      \\
                                                & Stack       & 6.61     & 1.5 & 75.37      \\
                                                & Interpolate & 6.61     & 1.5 & 75.39      \\
                                                & SWA         & 6.61     & 1.5 &  75.23     \\
                                                & ScaleNet    & 6.45     & 2.6 & 76.46      \\
        \midrule
        \multirow{6}{*}{Deit-Small~\cite{deit}} & -           & 22.05    & 4.6 & 79.86      \\
                                                & Random      & 23.83    & 5.0 & 80.07      \\
                                                & Stack       & 23.83    & 5.0 & 80.09      \\
                                                & Interpolate & 23.83    & 5.0 & 80.22      \\
                                                & SWA         & 23.83    & 5.0 & 80.09      \\
                                                & ScaleNet    & 23.53    & 9.2 & 81.13      \\
        \midrule
        \multirow{6}{*}{Deit-Base~\cite{deit}}  & -           & 86.57    & 17.6 & 81.80      \\
                                                & Random      & 93.66    & 19.0 & 82.22      \\
                                                & Stack       & 93.66    & 19.0 & 82.27      \\
                                                & Interpolate & 93.66    & 19.0 & 82.27      \\
                                                & SWA         & 93.66    & 19.0 & 82.34      \\
                                                & ScaleNet    & 89.55    & 35.2 & 82.53      \\
        \midrule
        \multirow{6}{*}{Swin-Tiny~\cite{swin}}  & -           & 28.29    & 4.5 & 81.20      \\
                                                & Random      & 30.06    & 4.8 & 81.16      \\
                                                & Stack       & 30.06    & 4.8 & 81.14      \\
                                                & Interpolate & 30.06    & 4.8 & 81.14      \\
                                                & SWA         & 30.06    & 4.8 & 81.28      \\
                                                & ScaleNet    & 29.37    & 5.2 & 81.43      \\
        \midrule
        \multirow{6}{*}{Swin-Small~\cite{swin}} & -           & 49.61    & 8.8 & 83.22      \\
                                                & Random      & 51.38    & 9.2 & 83.18      \\
                                                & Stack       & 51.38    & 9.2 & 83.07      \\
                                                & Interpolate & 51.38    & 9.2 & 83.05      \\
                                                & SWA         & 51.38    & 9.2 & 83.18      \\
                                                & ScaleNet    & 51.30    & 9.5 & 83.34      \\
        \midrule
        \multirow{6}{*}{Swin-Base~\cite{swin}}  & -           & 87.77    & 15.5 & 83.50      \\
                                                & Random      & 90.92    & 16.1 & 83.57      \\
                                                & Stack       & 90.92    & 16.1 & 83.53      \\
                                                & Interpolate & 90.92    & 16.1 & 83.52      \\
                                                & SWA         & 90.92    & 16.1 & 83.63      \\
                                                & ScaleNet    & 90.03    & 16.7 & 83.69      \\
        \bottomrule
    \end{tabular}}
    \label{tab:main}
\end{table}

\subsubsection{Optimization}

When tuning each scaled model, we use AdamW~\cite{adamw} as the optimizer, using an initial learning rate of 2e-4 and a cosine learning rate schedule, with 2e-6 as the final learning rate. Each model is trained for 100 epochs with a global batch size of 512, using 8 GPUs. The data augmentation configurations follow the official training script of Swin~\cite{swin}, except for the drop path ratio, which we have found to play an important role in affecting the performance of scaled models. In general, a significantly larger drop path ratio should be adopted for tuning scaled models. In our ablation study, we study the effect of this factor in more detail.

\subsubsection{Baselines}

To compare the ScaleNet method with existing approaches, we mainly consider the following two baselines:

\changed{\margintextnote{R1-Q2}\textbf{Random.} This baseline inserts new layers at the tail of the model and initializes their parameters randomly. It serves as the simplest model scaling method.}

\textbf{Stack.} This baseline achieves model scaling by stacking new layers at the tail of the pretrained model, with each layer initialized following the rule of StackBERT~\cite{gong2019efficient} in Equation~\ref{eq:baseline}.

\textbf{Interpolation.} This baseline interleaves new layers into the pretrained model and initializes them using weights from adjacent pretrained layers, following the rule of Interpolation in Equation~\ref{eq:baseline}. Instances of this baseline include~\cite{chang2017multi,dong2020towards}.

\changed{\textbf{Stochastic weight averaging (SWA).} This baseline inserts new layers at the tail of the model, with parameters initialized by averaging randomly sampled layers. SWA was originally proposed to average multiple checkpoints during model training.}

Since weight sharing is not adopted in either of these two baseline methods, to ensure a fair comparison, we reduce the number of newly added layers in both baselines to keep the total parameter count of the scaled model similar to that of ScaleNet. This allows us to demonstrate the parameter efficiency of ScaleNet. Specifically, this is achieved by adding layers one by one until the scaled model reaches a parameter count comparable to ScaleNet.

\subsection{Results on ImageNet-1K}

\begin{changedblock}
We begin our experiments with the ImageNet dataset~\cite{imagenet}, comparing ScaleNet with the following baselines: Random, Stack, Interpolate, and SWA. The results are presented in Table~\ref{tab:main}, which reports Top-1 accuracy alongside the number of parameters and computational complexity (FLOPs).
The table highlights the trade-offs inherent in model scaling. While doubling a model depth necessarily doubles its inference FLOPs, strength of ScaleNet lies in its exceptional parameter and training efficiency. Our method achieves superior performance with only a marginal increase in unique parameters compared to the original model, minimizing storage overhead. Furthermore, this performance is achieved at a fraction of the training cost required to train a similarly-sized model from scratch, which will be shown in Section~\ref{sec:exp:efficienct}.\margintextnote{\makecell{R1-Q2\\R2-Q1}}

Across all model variants, ScaleNet consistently outperforms the baselines. For example, the scaled DeiT-Base model reaches 82.53\% Top-1 accuracy, surpassing the strongest baseline while maintaining the lowest parameter overhead among expansion methods. This pattern holds for other models like Swin-Base, where ScaleNet achieves 83.69\% accuracy. This demonstrates that for a given computing budget, ScaleNet provides a more effective method for leveraging pretrained knowledge to build more powerful models.
\end{changedblock}

Overall, ScaleNet provides more efficient scaling by improving performance without significantly increasing the number of parameters, making it a cost-effective solution for scaling pretrained ViT models.

\begin{figure}
    \centering
    \includegraphics[width=0.95\linewidth]{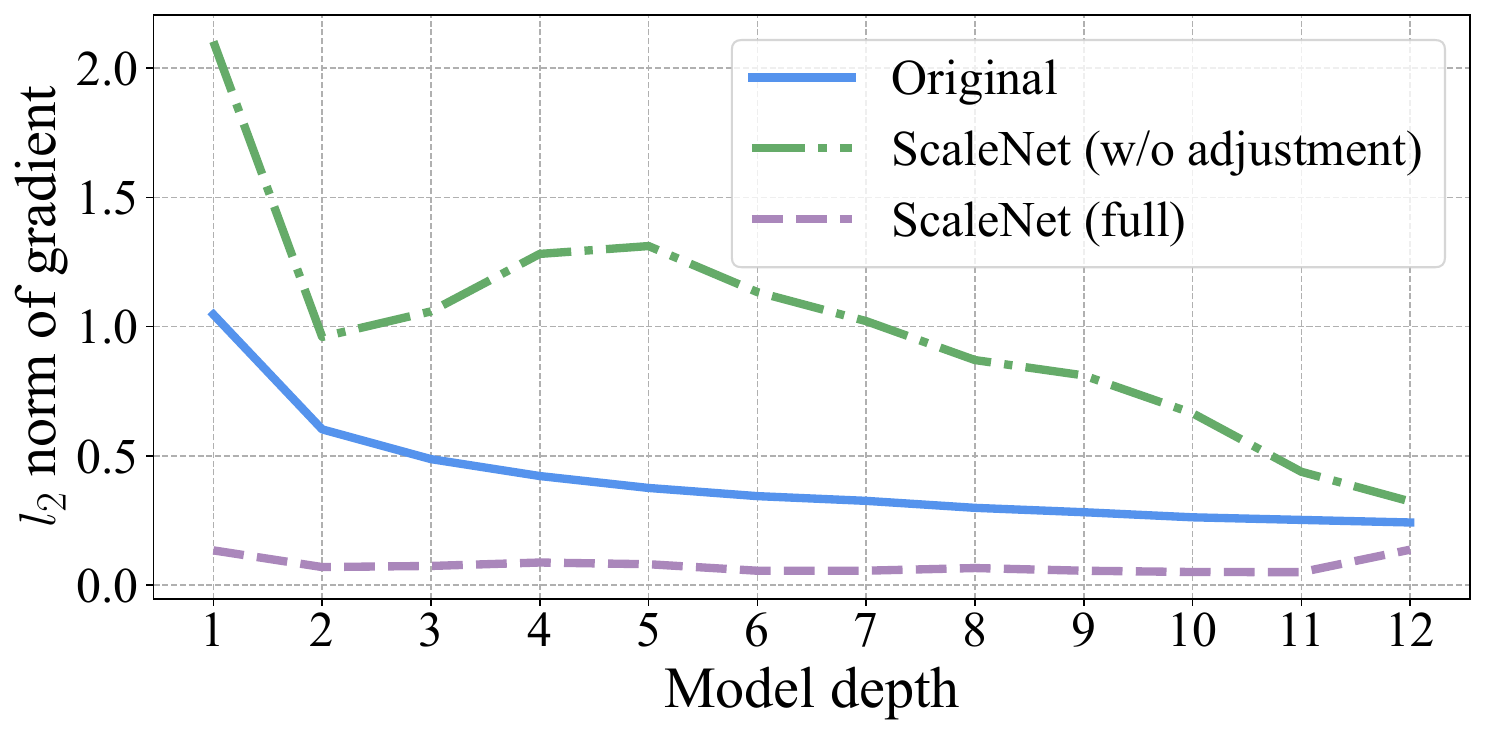}
    \caption{Comparison of the $l_2$ norm of gradients for each layer in the DeiT-B model scaled using different methods. Our method presents more stable gradients during training.}
    \label{fig:grad}
\end{figure}

\begin{figure*}
    \centering
    \begin{subfigure}{0.99\linewidth}
        \includegraphics[width=\textwidth]{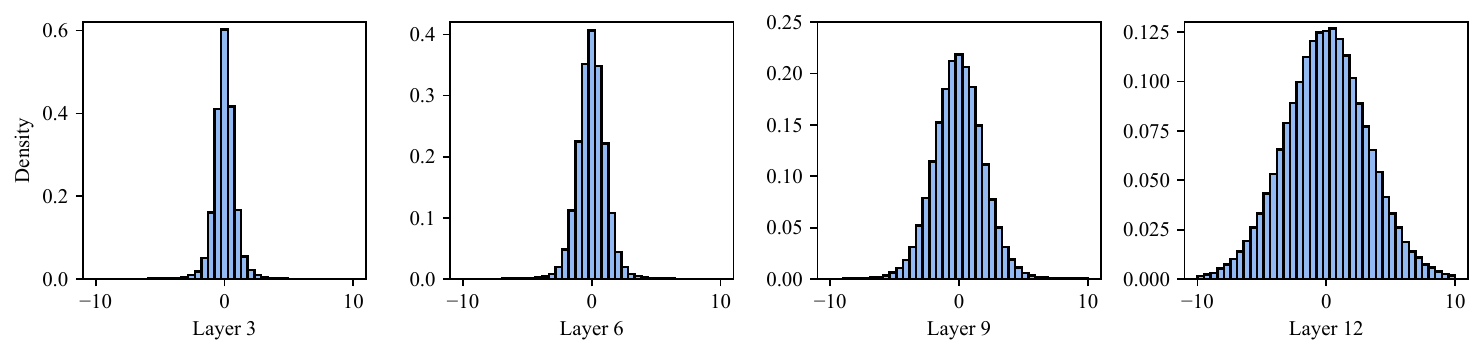}
        \caption{Original pretrained model (Top-1: 81.80\%)}
        \label{fig:hist_1}
    \end{subfigure} \\[1em]
    \begin{subfigure}{0.99\linewidth}
        \includegraphics[width=\textwidth]{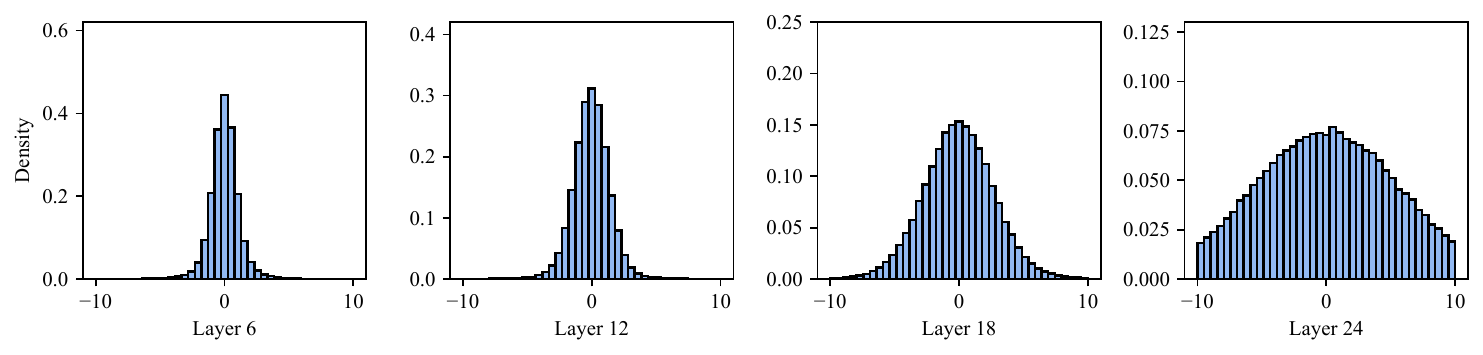}
        \caption{ScaleNet before finetuning (Top-1: 65.80\%)}
        \label{fig:hist_2}
    \end{subfigure} \\[1em]
    \begin{subfigure}{0.99\linewidth}
        \includegraphics[width=\textwidth]{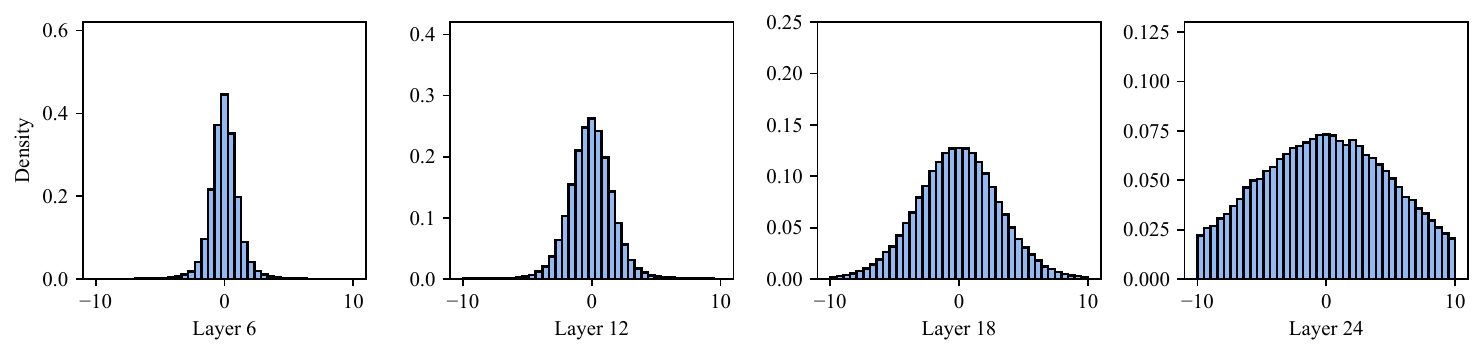}
        \caption{ScaleNet after finetuning (Top-1: 82.53\%)}
        \label{fig:hist_3}
    \end{subfigure} \\
    \caption{Feature distribution comparison between the original pretrained DeiT-Base model, ScaleNet before training, and ScaleNet after training. The histograms show the activation distributions for various layers, illustrating how the feature distribution evolves as the model is scaled and fine-tuned. The density in each figure on the y-axis is scaled to the same scale, and the x-axis is truncated to the range $[-10, 10]$.}
    \label{fig:hist}
\end{figure*}

\subsubsection{Analysis of gradient}

Previous work~\cite{minivit} has pointed out that the application of weight sharing can cause unstable gradients across model layers, which may negatively impact performance. To investigate the impact of weight sharing in our proposed method, we analyze the gradient norms at each layer of the model. Specifically, we use DeiT-Base for our analysis and compare three variations: the original model, a scaled model without layer adjustment, and a model scaled using the full ScaleNet approach.

We compute the gradient for each model using a batch of 256 samples and calculate the $l_2$ norm of the gradient in each layer for comparison. The results are presented in Figure~\ref{fig:grad}, where the gradient norm of the scaled model is computed over each scaled layer. Hence, it only presents 12 independent layers in this figure.

From the results, the original model shows a trend of gradually decreasing gradient norms as the model depth increases. However, when layer adjustment is not applied to the model, the gradient norms increase significantly, with fluctuations observed in the initial layers. In contrast, our ScaleNet approach results in stable gradients across the entire model, with norm values lower than those of the other two model variants. This demonstrates that weight sharing-based progressive training and layer adjustment mechanisms in ScaleNet effectively mitigate the instability caused by weight sharing, ensuring more stable and consistent gradient flow throughout the model.

\subsubsection{Analysis of feature}

To investigate how the scaling process affects the model, we analyze the distribution of intermediate features and their relationships across different layers. Specifically, we adopt the DeiT-Base model and compare features of the original pretrained model and the scaled model from two aspects: feature distribution and feature similarity.

\textbf{Feature distribution.}
We conduct an analysis of the feature distribution across three different versions of the DeiT-Base model: the original pretrained model, ScaleNet before training, and ScaleNet after training. Specifically, we provide histograms of the activations from uniformly sampled layers in these models. This allows us to visualize how the distribution of features evolves throughout the scaling process and training.

The histograms for the activations of selected layers are shown in Figure~\ref{fig:hist}. In the original pretrained model (Figure~\ref{fig:hist_1}), the feature distributions across different layers exhibit typical Gaussian-like distributions with a relatively narrow spread around zero. For ScaleNet before training (Figure~\ref{fig:hist_2}), the distributions show a similar pattern, although slightly broader compared to the original model. This is likely due to the additional layers introduced by ScaleNet, which may alter the feature distribution to some extent.

After training the ScaleNet model (Figure~\ref{fig:hist_3}), the distributions appear similar to the original model but with slightly wider spreads, indicating that training has allowed the model to adjust the feature distributions for better task-specific performance. Overall, the distributions in ScaleNet before and after training suggest that the scaling process does not drastically change the nature of the learned features but rather fine-tunes them for the new architecture.

\begin{figure*}
    \centering
    \includegraphics[width=0.8\linewidth]{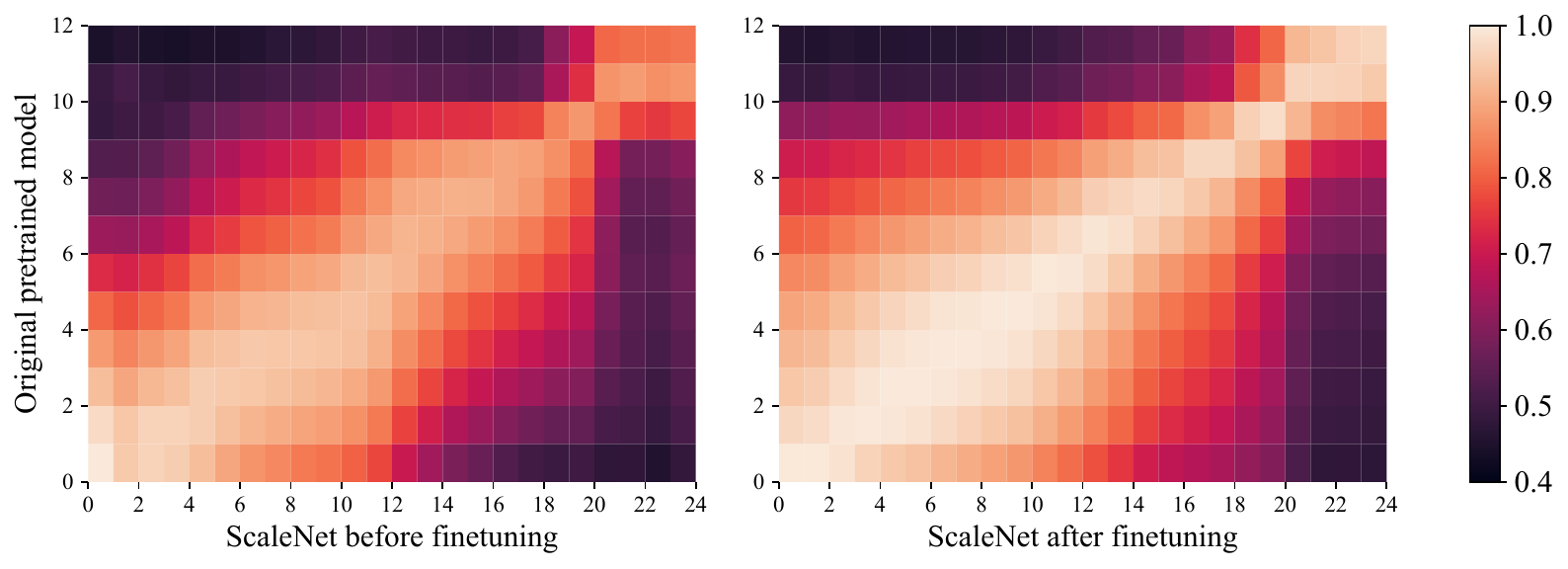}
    \caption{CKA similarity analysis between the original pretrained DeiT-Base model and ScaleNet before and after finetuning. The heatmaps show the alignment of feature representations across layers, with values closer to 1 indicating higher similarity. The results demonstrate that the scaled model before finetuning is already closely aligned with the original model, and finetuning further enhances this alignment. Coordinate axes indicate the layer indices.}
    \label{fig:cka}
\end{figure*}

\textbf{Feature similarity.}
To further understand the effects of ScaleNet on the feature representations, we perform a center kernel analysis (CKA)~\cite{cka} between the features of the original pretrained model and the features from ScaleNet after training. CKA is a measure of similarity that quantifies how similar the representations are between two models or layers, allowing us to determine if the newly added layers in ScaleNet align with the original pretrained model in terms of learned features.

In the left heatmap, which shows the CKA similarity before finetuning, we observe that the similarity across the layers is quite high, indicating that the scaled model closely mirrors the feature representations of the original model, but when we increase distance between layers, the alignment between layers decreases. This indicates that before training, new layers introduced by ScaleNet still maintain a strong alignment with the original pretrained model, though some divergence occurs.

After finetuning, we see an even higher alignment between the two models. The CKA similarity further increases, indicating that the finetuning process effectively brings the new layers into closer alignment with the representations of the original model. This shows that ScaleNet maintains the key features of the pretrained model while adapting the newly inserted layers to fit better with the overall model after training.

\subsection{Ablation Study}

\subsubsection{Effectiveness of each module}

\begin{changedblock}
\margintextnote{R2-Q1}
In this section, we present a comprehensive evaluation of the key components of ScaleNet, namely the weight sharing-based progressive training (WSPT) and the adjustment module. For experiments adopting adjustment modules without WSPT, we adjust the intermediate dimension of the adjustment modules to ensure that the overall parameter count of these models remains similar to that of ScaleNet.

Table~\ref{tab:ablation:main} presents the ablation results. From the results, WSPT consistently improves performance across most models, demonstrating its effectiveness as a strategy for scaling pretrained vision transformers. For example, in DeiT-Tiny, enabling WSPT without parameter expansion increases Top-1 accuracy from 72.16\% to 75.58\%, and similar improvements are observed in larger models like DeiT-Base and Swin-Base. However, it is worth noting that in some cases, such as Swin-Small, WSPT alone results in a slight performance drop, suggesting that the benefits of WSPT may vary depending on the model architecture and size. 
The introduction adjustment module also plays a critical role in improving performance. Adapter-based layer adjustment generally outperforms LoRA, achieving higher Top-1 accuracy in most cases. However, in results of Swin models, LoRA expansion results in a significant performance drop compared to the original pretrained model. 
When combining Adapter-based adjustment with WSPT, it delivers the best performance across all models. For example, in DeiT-Tiny, this combination achieves a Top-1 accuracy of 76.46\%, significantly higher than the baseline and other configurations. Similarly, in Swin-Base, the combination improves accuracy from 83.50\% to 83.69\%. This demonstrates that Adapter-based adjustment, when paired with WSPT, provides a robust mechanism for scaling pretrained models effectively, validating ScaleNet as an effective approach for scaling pretrained models.

\begin{table}
    \renewcommand\tabcolsep{7pt}
    \centering
    \caption{\changed{\intextnote{R2-Q1}Ablation study evaluating the impact of layer adjustment and WSPT on performance. The combination of Adapter-based layer adjustment and WSPT consistently delivers the best performance across all models. WSPT: weight sharing-based progressive training.}}
    \changedtab{\begin{tabular}{l|c|c|c|c}
        \toprule
        Model                                   & WSPT     & Adjustment   & \#Params (M) &  Top-1 \\
        \midrule
        \multirow{5}{*}{Deit-Tiny~\cite{deit}}  & -      & -        & 5.72     & 72.16      \\
                                                & \cmark & -        & 5.72     & 75.58      \\
                                                & \xmark & LoRA     & 6.31     & 74.86      \\
                                                & \xmark & Adapter  & 6.45     & 75.07      \\
                                                & \cmark & Adapter  & 6.45     & 76.46      \\
        \midrule
        \multirow{5}{*}{Deit-Small~\cite{deit}} & -      & -        & 22.05    & 79.86      \\
                                                & \cmark & -        & 22.05    & 80.56      \\
                                                & \xmark & LoRA     & 23.23    & 80.02      \\
                                                & \xmark & Adapter  & 23.53    & 80.01      \\
                                                & \cmark & Adapter  & 23.53    & 81.13      \\
        \midrule
        \multirow{5}{*}{Deit-Base~\cite{deit}}  & -      & -        & 86.57    & 81.80      \\
                                                & \cmark & -        & 86.57    & 82.08      \\
                                                & \xmark & LoRA     & 88.93    & 82.19      \\
                                                & \xmark & Adapter  & 89.55    & 82.28      \\
                                                & \cmark & Adapter  & 89.55    & 82.53      \\
        \midrule
        \multirow{5}{*}{Swin-Tiny~\cite{swin}}  & -      & -        & 28.29    & 81.20      \\
                                                & \cmark & -        & 28.29    & 81.23      \\
                                                & \xmark & LoRA     & 28.85    & 80.95      \\
                                                & \xmark & Adapter  & 28.99    & 81.32      \\
                                                & \cmark & Adapter  & 29.37    & 81.43      \\
        \midrule
        \multirow{5}{*}{Swin-Small~\cite{swin}} & -      & -        & 49.61    & 83.22      \\
                                                & \cmark & -        & 49.61    & 83.09      \\
                                                & \xmark & LoRA     & 50.76    & 82.93      \\
                                                & \xmark & Adapter  & 51.05    & 83.24      \\
                                                & \cmark & Adapter  & 51.30    & 83.34      \\
        \midrule
        \multirow{5}{*}{Swin-Base~\cite{swin}}  & -      & -        & 87.77    & 83.50      \\
                                                & \cmark & -        & 87.77    & 83.61      \\
                                                & \xmark & LoRA     & 89.31    & 83.26      \\
                                                & \xmark & Adapter  & 89.69    & 83.56      \\
                                                & \cmark & Adapter  & 90.03    & 83.69      \\
        \bottomrule
    \end{tabular}}
    \label{tab:ablation:main}
\end{table}

\end{changedblock}

\subsubsection{Weight sharing-based progressive training}

\begin{table}
    \renewcommand\tabcolsep{12pt}
    \centering
    \caption{Ablation study evaluating the impact of location of shared layers.}
    \begin{tabular}{l|c|c|c}
        \toprule
        Model                                  & Shared layers & \#Params (M) & Top-1 \\
        \midrule
        \multirow{6}{*}{Deit-Base~\cite{deit}} & -             & 86.57    & 81.80      \\
                                               & \{1,2,3\}     & 87.31    & 82.11      \\
                                               & \{4,5,6\}     & 87.31    & 82.17      \\
                                               & \{7,8,9\}     & 87.31    & 82.13      \\
                                               & \{10,11,12\}  & 87.31    & 82.30      \\
                                               & all           & 89.55    & 82.53      \\
        \midrule
        \multirow{6}{*}{Swin-Base~\cite{swin}} & -             & 87.77    & 83.50      \\
                                               & stage1        & 89.74    & 83.56      \\
                                               & stage2        & 89.78    & 83.63      \\
                                               & stage3        & 91.20    & 83.46      \\
                                               & stage4        & 90.03    & 83.69      \\
                                               & all           & 91.67    & 83.48      \\
        \bottomrule
    \end{tabular}
    \label{tab:ablation:layer}
\end{table}

\changed{\margintextnote{R1-Q4}The choice of which layers to reuse in weight sharing-based progressive training significantly impacts the performance of scaled models. To study this effect, we conduct experiments comparing different sets of reused layers for DeiT-Base and Swin-Base, with the results provided in Table~\ref{tab:ablation:layer}. For DeiT-Base, which has an isotropic architecture where all layers are structurally identical and operate on features of a constant resolution, sharing parameters across all layers proves to be the most effective strategy. As the table shows, performance peaks at an 82.53\% Top-1 accuracy when all layers are included for reuse. This suggests that the homogeneity of isotropic models makes their layers highly interchangeable and well-suited for comprehensive sharing. In contrast, Swin-Base employs a hierarchical architecture composed of distinct stages that process feature maps at varying spatial resolutions and channel dimensions. For such models, sharing layers across these functionally specialized stages is detrimental, as it forces layers designed for different scales to use the same weights. This architectural conflict explains why reusing all layers degrades performance to 83.48\%. The optimal strategy is therefore to limit sharing to layers within a single, homogeneous stage. Our results validate this, as reusing only the layers in the stage4 achieves the best performance of 83.69\%.}

\begin{figure*}
    \centering
    \begin{subfigure}{0.48\linewidth}
        \includegraphics[width=\textwidth]{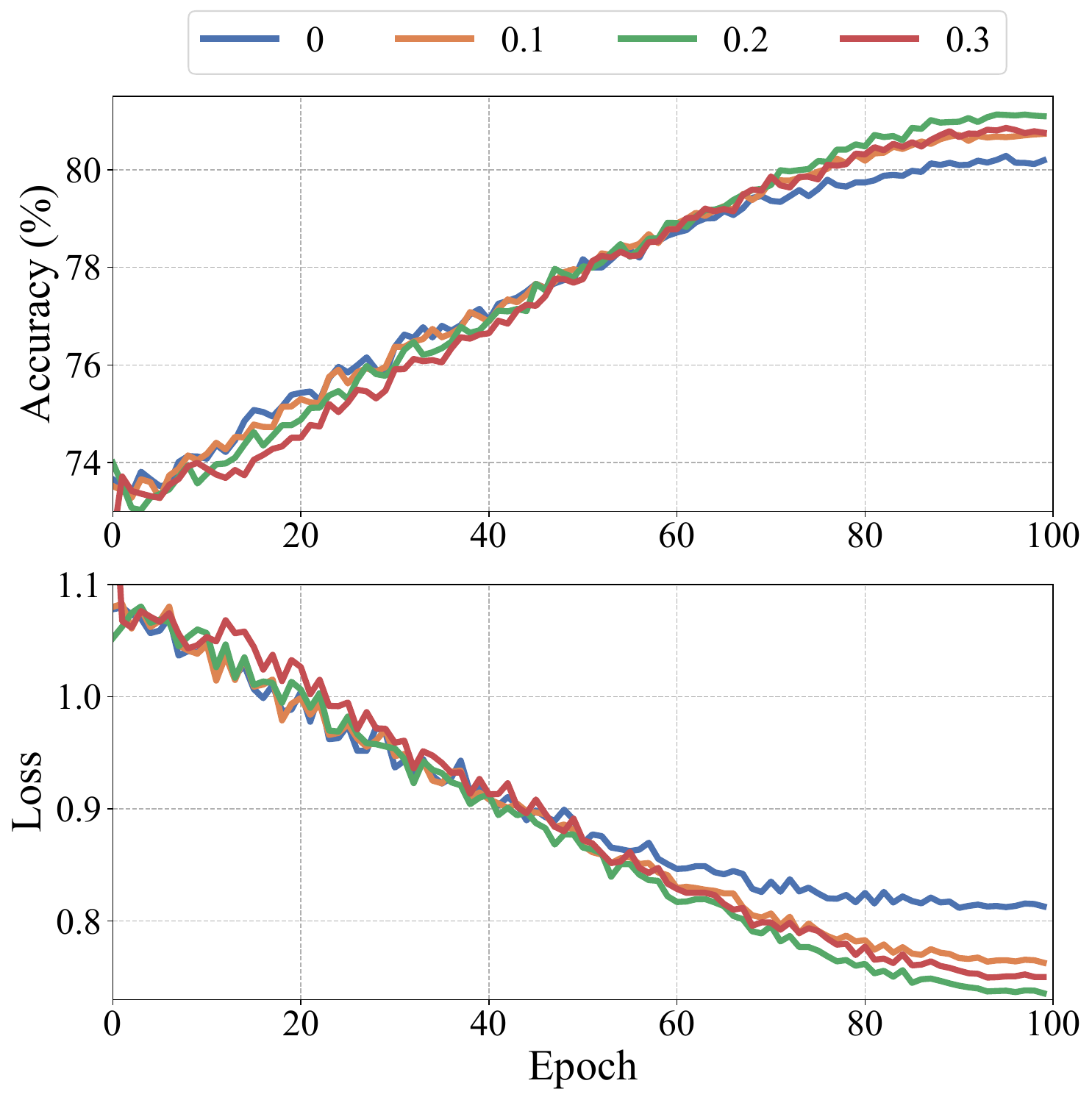}
        \caption{DeiT-Small}
        \label{fig:dp_1}
    \end{subfigure}
    \hfill
    \begin{subfigure}{0.48\linewidth}
        \includegraphics[width=\textwidth]{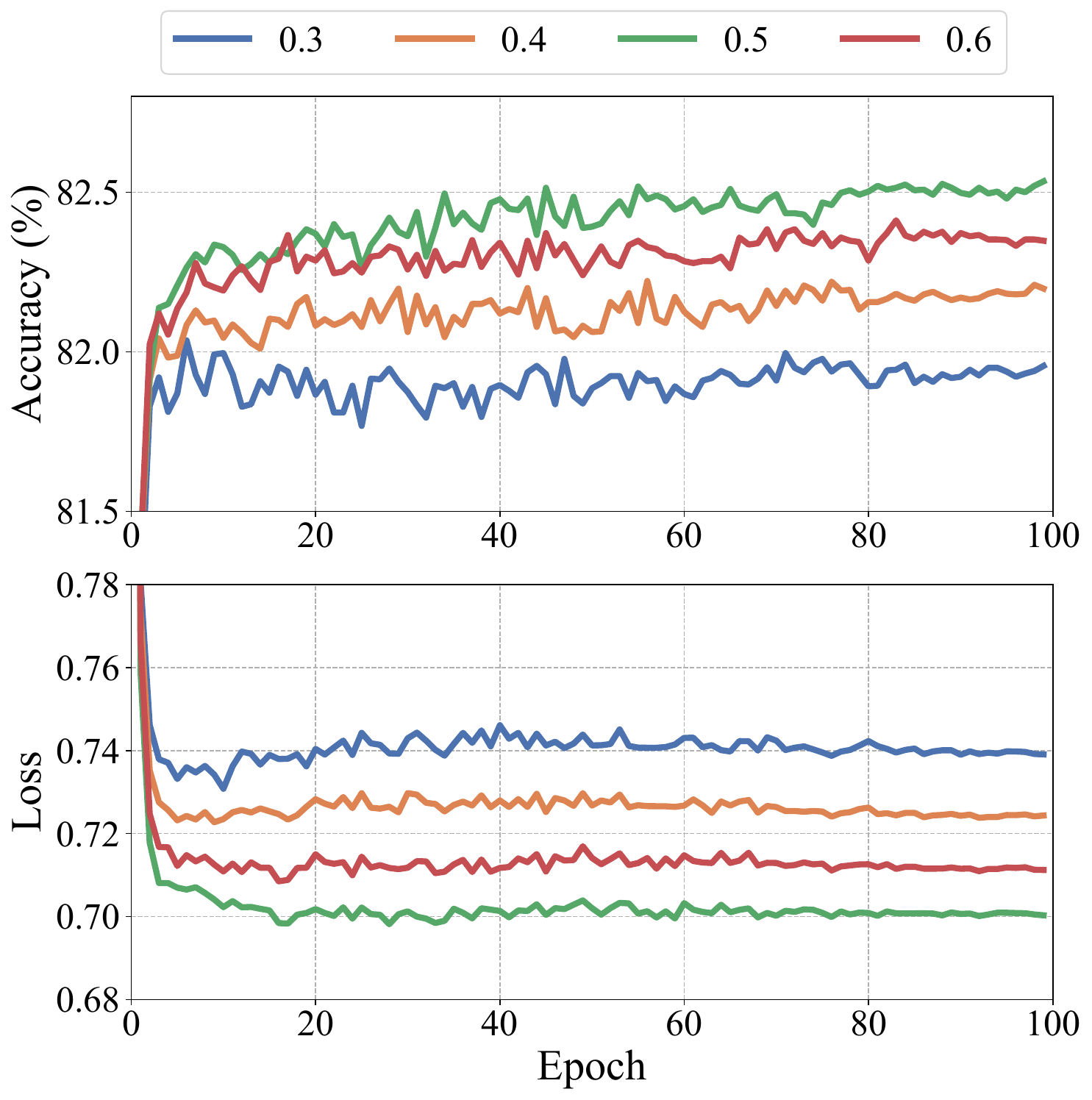}
        \caption{DeiT-Base}
        \label{fig:dp_2}
    \end{subfigure}
    \caption{Effect of different drop path ratios on the performance of DeiT-Small and DeiT-Base models.}
    \label{fig:dp}
\end{figure*}

\begin{table}
    \renewcommand\tabcolsep{12pt}
    \centering
    \caption{Impact of layer reuse times on DeiT-Base and Swin-Base models.}
    \begin{tabular}{l|c|c|c}
        \toprule
        Model                                  & Shared times & \#Params (M) & Top-1 \\
        \midrule
        \multirow{3}{*}{Deit-Base~\cite{deit}} & -            & 86.57    & 81.80      \\
                                               & 2            & 89.55    & 82.53      \\
                                               & 3            & 92.53    & 82.27      \\
        \midrule
        \multirow{3}{*}{Swin-Base~\cite{swin}} & -            & 87.77    & 83.50      \\
                                               & 2            & 90.03    & 83.69      \\
                                               & 3            & 91.39    & 83.58      \\
        \bottomrule
    \end{tabular}
    \label{tab:ablation:time}
\end{table}

To further investigate the impact of weight sharing-based progressive training, we study how the number of times layers are reused affects model performance. Table~\ref{tab:ablation:time} presents the results of this ablation, comparing different reuse times for DeiT-Base and Swin-Base. The results reveal that reusing layers twice improves performance for both models, with DeiT-Base achieving a Top-1 accuracy of 82.53\% and Swin-Base reaching 83.69\%. However, increasing the reuse time to three leads to a performance drop, with DeiT-Base dropping to 82.27\% and Swin-Base to 83.58\%. This suggests that reusing the same parameters too many times can constrain ability of the model, likely due to over-reliance on shared weights. highlighting the importance of balancing reuse frequency to avoid diminishing returns in model performance.

\begin{table}
    \renewcommand\tabcolsep{12pt}
    \centering
    \caption{Impact of layer organization strategies on DeiT-Base and Swin-Base models.}
    \begin{tabular}{l|c|c}
      \toprule
      Model                                  & Layer order & Top-1 \\
      \midrule
      \multirow{3}{*}{Deit-Base~\cite{deit}} & -           & 81.80      \\
                                             & Interpolate & 82.53      \\
                                             & Stack       & 82.23      \\
      \midrule
      \multirow{3}{*}{Swin-Base~\cite{swin}} & -           & 83.50      \\
                                             & Interpolate & 83.51      \\
                                             & Stack       & 83.69      \\
      \bottomrule
    \end{tabular}
    \label{tab:ablation:order}
  \end{table}

In addition to reuse time, we explore how the organization of reused layers impacts model performance. Specifically, we compare the two strategies presented in Equation~\ref{eq:baseline}. For Swin models with different input sizes at each stage, we perform weight sharing-based progressive training at the stage level. Table~\ref{tab:ablation:order} summarizes the results. For DeiT-Base, interpolating layers yields the best performance, achieving a Top-1 accuracy of 82.53\%, compared to 82.23\% for stacking layers, indicating that inserting layers within the pretrained model allows for better integration of new and existing features. In contrast, for Swin-Base, stacking layers at the tail achieves the highest accuracy. This difference may stem from the distinct architectures of DeiT and Swin models, which are isotropic and hierarchical, respectively.

\subsubsection{Layer adjustment}

\begin{table}
    \renewcommand\tabcolsep{12pt}
    \centering
    \caption{Impact of layer adjustment designs on DeiT-Base.}
    \begin{tabular}{l|c|c|c}
        \toprule
        Adjustment & Middle dim & \#Params (M) & Top-1 \\
        \midrule
        -             & -          & 86.57    & 81.80      \\
        LoRA          & 4          & 87.75    & 82.15      \\
        LoRA          & 8          & 88.93    & 82.17      \\
        LoRA          & 16         & 91.29    & 82.18      \\
        Adapter       & 8          & 88.06    & 82.37      \\
        Adapter       & 16         & 89.55    & 82.53      \\
        Adapter       & 32         & 92.53    & 82.44      \\
        \bottomrule
    \end{tabular}
    \label{tab:ablation:delta}
\end{table}

In this section, we investigate the impact of different layer adjustment designs on model performance when adopting weight sharing-based progressive training. We compare LoRA and Adapter as adjustment modules with varying intermediate dimensions, using DeiT-Base as the test model. The results are presented in Table~\ref{tab:ablation:delta}.

The experiments demonstrate that both LoRA and Adapter adjustments improve performance compared to the baseline. Adapter-based adjustments consistently outperform LoRA, achieving the highest Top-1 accuracy of 82.53\%. This demonstrates that adapters provide a more effective mechanism for enhancing the capacity of shared layers. Interestingly, increasing the intermediate dimension for adapters beyond, \eg, from 16 to 32, results in a slight performance drop. This suggests that while larger intermediate dimensions can enhance capacity, they may also introduce redundancy, preventing the reuse of pretrained knowledge and thus leading to suboptimal performance.

\subsubsection{Drop path}

The drop path hyperparameter plays a pivotal role in determining the performance of ScaleNet. To study its effect, we compared different drop path value configurations using DeiT-Small and DeiT-Base models. Our results, as shown in Figure~\ref{fig:dp}, reveal that a higher drop path ratio leads to improved performance for both models in terms of accuracy and loss reduction. For instance, when adopting a drop path ratio of 0.2, ScaleNet with DeiT-Small achieves approximately a 1\% improvement in accuracy and a 0.1 reduction in loss. This suggests that for scaled models, a larger drop path ratio should be favored during training, as it helps the model generalize better by preventing overfitting. However, further increases in the drop path ratio beyond this point lead to performance degradation.
For the larger DeiT-Base model, where only a small subset of parameters is trainable, the effect of varying the drop path ratio holds, with the optimal configuration being a ratio of 0.5. In our experiments, we set the drop path ratio to 0.5 for DeiT-Base and Swin-Base models, and 0.2 for all other models, except for DeiT-Tiny, where the drop path ratio is set to 0.

\subsection{Training Efficiency}
\label{sec:exp:efficienct}

To demonstrate the efficiency of ScaleNet, we compare its performance with training scaled models from scratch without weight sharing-based progressive training. We utilized both the DeiT-Small and DeiT-Base models, each scaled to 24 layers. The pretraining process for both models followed the official implementation of DeiT, with training period of 100 and 300 epochs.

The results shown in Table 1 clearly highlight the efficiency of ScaleNet in terms of both training time and performance. For DeiT-Small, training from scratch for 100 epochs took 22.5 hours, achieving a top-1 accuracy of 77.32\%. When pretrained for 300 epochs, the model's accuracy increased to 79.31\%, but this came at a significantly higher cost of 47.3 hours. In contrast, using ScaleNet with DeiT-Small for just 100 epochs achieved a top-1 accuracy of 81.13\%, surpassing the accuracy of the fully pretrained model, while also reducing the training time to just 15.8 hours. Similarly, this trend is also observed for DeiT-Base. These results demonstrate that increasing model depth during pretraining can impede the training process, leading to longer training times and, in some instances, lower performance. By leveraging the pretrained model and efficiently expanding its architecture, ScaleNet provides a more effective alternative—significantly reducing both training time and computational resources while improving model performance.

\begin{table}
    \renewcommand\tabcolsep{6pt}
    \centering
    \caption{Comparison of training time and performance between pretraining without weight sharing-based progressive training, and ScaleNet for scaled models.}
    \begin{tabular}{l|c|c|c|c}
      \toprule
      Model & Method   & Epochs & Time (h) & Top-1 \\
      \midrule
      \multirow{3}{*}{\makecell{Deit-Small~\cite{deit}  \\(24 layers)}}  & Pretrain &100 & 22.5 & 77.32      \\
            & Pretrain & 300    & 47.3     & 79.31      \\
            & ScaleNet & 100    & 15.8     & 81.13      \\
      \midrule
      \multirow{3}{*}{\makecell{Deit-Base~\cite{deit}   \\(24 layers)}}  & Pretrain &100 & 29.8 & 68.91      \\
            & Pretrain & 300    & 89.5     & 75.11      \\
            & ScaleNet & 100    & 19.8     & 82.53      \\
      \bottomrule
    \end{tabular}
    \label{tab:ablation:speed}
  \end{table}

\subsection{Extending to More Tasks}

\begin{table}
    \centering
    \caption{Comparison of pretrained and scaled models on COCO object detection task.}
    \begin{tabular}{l|c|c|ccc}
        \toprule
        Model                                   & Expand   & \#Params (M) & AP   & AP$_{50}$ & AP$_{75}$ \\
        \midrule
        \multirow{2}{*}{Swin-Tiny~\cite{swin}}  & -        & 47.8     & 46.0 & 68.1      & 50.6      \\
                                                & ScaleNet & 48.9     & 46.7 & 68.7      & 51.2      \\
        \midrule
        \multirow{2}{*}{Swin-Small~\cite{swin}} & -        & 69.1     & 48.2 & 69.8      & 52.8      \\
                                                & ScaleNet & 70.8     & 48.4 & 70.1      & 53.1      \\
        \bottomrule
    \end{tabular}
    \label{tab:det}
\end{table}

\subsubsection{Object detection}

To further evaluate the effectiveness of ScaleNet, we conduct experiments on object detection using the COCO 2017 dataset~\cite{coco} under the MMDetection framework~\cite{mmdetection}. Specifically, we adopt pretrained Mask R-CNN~\cite{he2017mask} models with Swin-Tiny (Swin-T) or Swin-Small (Swin-S) as the backbone for scaling. We include all layers for reuse, and other scaling setup for the backbones follows the same configuration as in the ImageNet-1K experiments. The scaled models are trained using a 1$\times$ schedule (12 epochs) with the default framework configuration. During training, the entire detection head is frozen except for the final classification and regression layers.

Table~\ref{tab:det} presents the results of the object detection experiments. With a fast adaptation process, the scaled detectors with both backbones achieve improved performance. Specifically, the average precision of the scaled detectors increases by 0.7\% for Swin-T and by 0.2\% for Swin-S. These improvements come with a negligible increase in parameters, demonstrating the efficiency of ScaleNet in detection task.

\begin{table}
    \centering
    \caption{\changed{\intextnote{R1-Q1/R3-Q2}Comparison of pretrained and scaled models on ADE20K semantic segmentation task.}}
    \changedtab{\begin{tabular}{l|c|c|ccc}
        \toprule
        Model                                   & Expand   & \#Params (M) & mIoU \\
        \midrule
        \multirow{2}{*}{Swin-Tiny~\cite{swin}}  & -        & 59.9     & 44.41     \\
                                                & ScaleNet & 63.1     & 44.69       \\
        \midrule
        \multirow{2}{*}{Swin-Small~\cite{swin}} & -        & 81.3     & 47.72       \\
                                                & ScaleNet & 84.4     & 48.13       \\
        \bottomrule
    \end{tabular}}
    \label{tab:seg}
\end{table}

\begin{changedblock}
    
\subsubsection{Semantic segmentation}

\margintextnote{\makecell{R1-Q1\\R3-Q2}}
We also evaluate ScaleNet on semantic segmentation task using ADE20K dataset~\cite{ade20k} using MMSegmentation framework~\cite{mmseg2020}. Similar to object detection experiments, we adopt Swin-T or Swin-S as the backbone for scaling under UPerNet~\cite{upernet}, training for 40K steps, with other configurations the same.

As shown in Table~\ref{tab:seg}, the segmentation results are consistent with those observed in object detection. When using Swin-Tiny and Swin-Small, the mIoU improves by 0.27 and 0.41, respectively. These results further demonstrate the effectiveness of ScaleNet for the segmentation task.

\subsubsection{Language modeling}

\margintextnote{\makecell{R1-Q1\\R2-Q2\\R3-Q2}}
Moreover, we conduct experiments on the language modeling task to evaluate the generalizability of ScaleNet. We adopt Llama-3.2-1B~\cite{grattafiori2024llama} as the pretrained model, using all layers to construct a 2$\times$ scaled model. The scaled model is trained for 3 epochs on the Stanford Alpaca dataset~\cite{alpaca}.
Zero-shot performance on common-sense reasoning benchmarks is used for evaluation, including BoolQ~\cite{clark2019boolq}, PIQA~\cite{bisk2020piqa}, HellaSwag~\cite{zellers2019hellaswag}, WinoGrande~\cite{sakaguchi2021winogrande}, ARC-easy~\cite{clark2018think}, ARC-challenge~\cite{clark2018think}, and OpenbookQA~\cite{mihaylov2018can}.

As shown in Table~\ref{tab:llm}, the scaled model outperforms the original model on almost all datasets, with the exception of OpenbookQA, where performance is slightly lower. Overall, the scaled model achieves an average performance improvement of 0.92\%. This experiment further demonstrates the generalizability of ScaleNet.

\end{changedblock}

\begin{table*}
    \centering
    \caption{\changed{\intextnote{R1-Q1/R2-Q2/R3-Q2}Comparison of pretrained and scaled models on language modeling tasks. The used LLM is Llama-3.2-1B.}}
    \changedtab{\begin{tabular}{c|c|ccccccc|c}
        \toprule
        Expand & \#Params (M) & BoolQ & PIQA & HellaSwag & WinoGrande & ARC-e & ARC-c & OpenbookQA & Average \\
        \midrule
        -      & 1236M        & 63.98 & 74.59& 63.66     & 60.69      & 60.48 & 36.26 & 37.20& 56.69   \\
        ScaleNet&1265M        & 66.12 & 75.73& 64.98     & 61.17      & 60.82 & 37.63 & 36.80& 57.61   \\
        \bottomrule
    \end{tabular}}
    \label{tab:llm}
\end{table*}

\section{Conclusion}
The escalating computational and environmental costs associated with scaling ViTs have underscored the need for parameter-efficient expansion strategies. In this work, we proposed ScaleNet, a novel framework that enables efficient scaling of pretrained ViT models through layer-wise weight sharing and lightweight adaptation. This approach addresses two critical limitations of existing techniques: the parameter inefficiency of naively duplicating or mapping layers during scaling, and the risk of performance degradation due to rigid weight sharing. By inserting additional layers that reuse pretrained parameters while incorporating minimal adjustment parameters via parallel adapter modules, ScaleNet achieves model expansion without the proportional increase in parameters. In our experiments on the ImageNet-1K dataset, ScaleNet improves the performance of the 2$\times$ depth-scaled DeiT-Base model by 7.42\% compared to scaling the model by training from scratch, while requiring only one-third of the training epochs. These results highlight the potential of ScaleNet to enhance the performance of ViTs while maintaining computational efficiency. 

The implications of this work extend beyond computational efficiency. By significantly reducing the storage and carbon footprint associated with training large ViTs, ScaleNet contributes to more sustainable AI development practices. Additionally, it democratizes access to high-performance vision models for researchers with limited computational resources. Future research directions include exploring dynamic weight-sharing mechanisms, extending the framework to hybrid architectures combining ViTs with convolutional networks, investigating the limits of scalability under increasingly constrained parameter budgets, and introducing knowledge distillation techniques~\cite{hinton2015distilling,hao2025toward} to further improve the performance. The principles established in ScaleNet may also inspire new paradigms for resource-efficient model adaptation in related domains such as natural language processing and multimodal learning.

% \section*{Acknowledgments}
% This should be a simple paragraph before the References to thank those individuals and institutions who have supported your work on this article.
% todo

\bibliographystyle{IEEEtran}
\bibliography{ref}

\end{document}